\documentclass[letterpaper, 10 pt, conference]{ieeeconf}  
\usepackage{graphicx}
\usepackage{caption}
\usepackage{xcolor}
\usepackage{subcaption}
\usepackage[pdftex,	
	    pdfstartview=FitH,
	    linkcolor=black,
	    citecolor=black,
	    urlcolor=black,
	    filecolor=black
	    ]{hyperref}
\usepackage{url}

\IEEEoverridecommandlockouts                              

\usepackage{lipsum}  

\title{\LARGE \bf
On Robot Acceptance and Trust: A Review and Unanswered Questions*
}

\author{Kerstin S. Haring$^{1}$, \textit{Member, IEEE}
\thanks{*This work was supported by the University of Denver's Professional Research Opportunities for Faculty (PROF) under grant \# 142101-84994}
\thanks{$^{1}$Kerstin S. Haring is with the Ritchie School of Computer Science and Engineering,
        University of Denver, Denver, CO 80210, USA,
        {\tt\small kerstin.haring@du.edu}}%
}

\begin{document}




\maketitle
\thispagestyle{empty}
\pagestyle{empty}



\begin{abstract}
This position paper briefly considers the current benefits and shortcomings surrounding robot trust and acceptance, focusing on robots with interactive capabilities. The paper concludes with currently unanswered questions and may serve as a jumping-off point for discussion. 
\end{abstract}

\section{Overview}
The acceptance of novel technologies and social robots has specifically been described as a critical factor to successfully deploy socially interactive robots on a large scale \cite{dahl2013robots, pani2020evaluating, whelan2018factors}. Robot acceptance is important for several reasons. For one, robots have the potential to improve the quality of life of individuals and communities through better health care, improved education, greater convenience, and more personalized experiences \cite{broadbent2011human, tapus2009improving}. Accepting robots enables humans (and potentially pets) to benefit from advances. 
Robots may provide an innovative solution to social problems, including climate change, resource scarcity, and education and healthcare disparities may also stimulate economic growth \cite{bednar2020socio, prassida2022conceptual}.
Robots can change the way we communicate and collaborate to be more efficient \cite{hayes2013challenges,breazeal2004tutelage} and provide access to information, tools, and resources, allowing people of various backgrounds to participate in economic, social, and educational activities, thus empowering people \cite{breazeal2004teaching,passey2018digital}. 
A drawback is that at the same time, robot acceptance may also have the opposite effect. The quality of life may be reduced by poorly designed robot systems that one may now depend on. Social inequalities may actually increase due to the high cost of robot systems and the lack of universal availability. Robots are also not unbiased machines and can reinforce gender stereotypes \cite{Perugia2023Models}, and their unjust use could deepen social injustices \cite{Winkle2023Feminist}.
More robots may be only an innovative solution for some individuals, and for others it means more resource scarcity and disparities. Similarly, as they may be economic growth for some industries, people fear that by accepting robots, they accept that their jobs are replaced by a machine. 

Trust in novel technologies and robots specifically has also been described as a critical factor in successfully enabling socially interactive robots to collaborate, interact and support humans \cite{sheridan2016human,haring2021applying,otting2022let,malle2020trust,nam2020trust,gao2013modeling}. When people trust a robot, they feel more confident in using it and are willing to take the risks associated with the adoption and acceptance of robot technologies. Even more so, with appropriately calibrated trust in robots, human task efficiency could increase significantly, as is the case in the field of automation \cite{ososky2013building,billings2012human, tenhundfeld2019calibrating}. When people trust a robot, they might display high confidence in the performance of the overall human-robot team or system and be able to deliver intended outcomes sooner and with higher confidence, promoting positive user experiences and high user satisfaction. Trustworthy robots also increase their market success and would be perceived as being developed and used responsibly, mitigating many of the ethical considerations and challenges outlined by the interactive robotics stakeholders \cite{darling2015s,darling2021new,haring2019dark}.

A huge drawback is that at the same time, trust towards robots may also have the opposite effect. When our trust in the robot is betrayed by the robot not showing the reliability, accuracy or functionality it was entrusted with, trust may quickly turn to distrust and disuse \cite{lewis2018role}, hard to repair \cite{baker2018toward}, and even may have dire consequences when a trusted system fails in real-world applications (see recent reports of increased accidents of autonomous vehicles). Humans do not take it lightly when a robot does not perform "as it should have performed". In addition, there may be less dire yet impactful consequences when users overtrust a robot when it is not appropriate, like a decrease in performance, breach of personal data and decrease in security, impeded decision making, and inappropriate risk mitigation. Ironically, the same goes for undertrusting a robot, resulting in underutilization of it's true capabilities with similar consequences. 

In addition, robot researchers and industry alike do not share a common definition of what they refer to when they talk about \textit{trust}. Without such a definition, it is difficult to develop robot trust metrics and interpret measurements from human-robot experiments or collected from robot users. The lack of a unifying definition has led to several research publications that redefine trust for their purpose and to explain research results. Although a researcher may be forced to do so when a clear definition is lacking, it makes it harder to compare trust between studies and experiments. At the same time, it is doubtful that the field of robotics can agree on one definition of \textit{robot trust} given that the term \textit{trust} itself is interpreted widely different by individuals. 





\section{Considerations}
When it comes to robot acceptance and robot trust, numerous questions remain unanswered. 
The first question is about reliability and safety. Is it actually possible to build robot systems that consistently perform tasks accurately and safely? While reliability is important for trust, unexpected errors can have serious consequences for trust and robot acceptance. If robot systems that are consistently accurate and safe are not possible, how do we design interactions for errors? Do we need to design for common errors rather than preventing them? How would we address unexpected errors when the robot has no "awareness" in a human sense of the error? 
The second question asks how ethical decision-making can be integrated into interactive robotic systems. Trust often requires the ability to navigate ethical dilemmas, an understanding of morals, and integrity. How would we present the integrity of the robot in the various systems and contexts of the robot?
The third question is unanswered not only for robots but also for current Artificial Intelligence (AI). How can robots and AI provide understandable explanations or transparency for their actions or decisions? Trust may be contingent on such transparency so people can comprehend robot behaviors (consider the question people may ask when observing a robot: "what is it doing now?"). The \textit{how} to present transparency and in what modality has not been answered conclusively. 
The fourth big question is how interactive robots handle privacy. Robots in our homes and workplaces, potentially present in some of our most private moments, raise the question if we can "trust" them with these personal moments, meaning that our data are secure and, on top of that, secure in the sense that we understand it to be secure, which does not mean the same for individuals or groups like families.
Lastly, we have yet to find an answer to design robots and interactions with robots that foster trust and clear, natural, accurate, and enjoyable communication. Such communications are likely to establish trust in their capabilities and intentions, but \textit{how} currently lacks guidelines. 
From a more philosophical approach, we must ask ourselves when a clear definition of trust and its metrics is lacking: Are we really looking at trust, or is it a multifaceted phenomenon that we need to break down into its facets \cite{Ullman2019Measuring}?  Would it be possible to achieve the goal of "trust" in robots with just finding an answer to some facets, and if yes, which ones?

\bibliographystyle{ieeetr}
\bibliography{references}

\end{document}